\documentclass[10pt,twocolumn,letterpaper]{article}

\usepackage[pagenumbers]{wacv} % To force page numbers, e.g. for an arXiv version

\usepackage{graphicx}
\usepackage{amsmath}
\usepackage{amssymb}
\usepackage{booktabs}

\usepackage{graphicx}
\usepackage{amsmath}
\usepackage{amssymb}
\usepackage{booktabs}
\usepackage{makecell}
\makeatletter
\@namedef{ver@everyshi.sty}{}
\makeatother
\usepackage{tikz}
\usetikzlibrary{tikzmark}
\usepackage{algorithm}
\usepackage{algpseudocode}
\usepackage{arydshln}
\usepackage{xcolor}         % colors
\usepackage{subfiles}
\usepackage{bm}
\usepackage{cite}
\usepackage{multirow}
\usepackage[accsupp]{axessibility} 
\usepackage{graphicx}
\usepackage{wrapfig}
\usepackage[font=small,labelfont=bf]{caption}
\usepackage{amsmath}
\usepackage{pifont}
\usepackage{colortbl}
\usepackage{enumitem}
\usepackage{multirow}
\usepackage{colortbl}
\usepackage{axessibility}

\usepackage[pagebackref,breaklinks,colorlinks]{hyperref}

\usepackage[capitalize]{cleveref}
\crefname{section}{Sec.}{Secs.}
\Crefname{section}{Section}{Sections}
\Crefname{table}{Table}{Tables}
\crefname{table}{Tab.}{Tabs.}

\def\wacvPaperID{2297} % *** Enter the WACV Paper ID here
\def\confName{WACV}
\def\confYear{2024}

\setlist{nolistsep}

\begin{document}

\newcommand{\tablebaselong}{
\begin{table*}[t]
    \centering
    \small
    \begin{tabular}{c|ccc|ccc|ccc|ccc}
    \hline
         \multicolumn{1}{c}{\multirow{2}{*}{ViT-B}}    & \multicolumn{3}{c}{$r=8$} & \multicolumn{3}{c}{$r=12$} & \multicolumn{3}{c}{$r=16$} & \multicolumn{3}{c}{$r=20$}  \\ \cline{2-13} 
         & Acc & Speed & FLOP & Acc & Speed & FLOP & Acc & Speed & FLOP & Acc & Speed & FLOP  \\  \hline
        Full Model & 83.74 & 323.61 & 17.58 & 83.74 & 323.61 & 17.58 & 83.74 & 323.61 & 17.58 & 83.74 & 323.61 & 17.58  \\ \hdashline[0.5pt/1pt]
        ToMe & 82.86 & 413.67 & 13.12 & 81.82 & 489.72 & 10.93 & 78.88 & 607.30 & 8.78 & 67.54 & 736.91 & 7.14  \\ 
        ToFu AVG & 83.19 & \textbf{417.74} & 13.12 & 82.43 & \textbf{494.41} & 10.93 & 80.43 & \textbf{615.53} & 8.78 & 72.19 & \textbf{748.75} & 7.14  \\ 
        ToFU MLERP & \textbf{83.22} & 413.86 & 13.12 & \textbf{82.46 }& 490.47 & 10.93 & \textbf{80.70} & 610.44 & 8.78 & \textbf{74.06} & 745.40 & 7.14 \\ \hline
    \end{tabular}
    
    \caption{Performance on ImageNet using ViT-B. As the reduction $r$ increases, the inference speed (in images/sec) rises, but at the expense of Top1 accuracy. ToFu AVG outperforms ToMe in both accuracy and speed, due to the hybrid merging. ToFu MLERP achieves better results than ToFu AVG with a slight speed reduction. The  trade-off between accuracy and speed favors ToFu MLERP, as shown in Fig.~\ref{fig:tradeoff}.}
    \label{tab:base}
\end{table*}

}

\newcommand{\tablelargelong}{
\begin{table*}[t]
    \centering
    \small
    \begin{tabular}{c|ccc|ccc|ccc|ccc}
    \hline
        \multicolumn{1}{c}{\multirow{2}{*}{ViT-L}}    & \multicolumn{3}{c}{$r=8$} & \multicolumn{3}{c}{$r=12$} & \multicolumn{3}{c}{$r=16$} & \multicolumn{3}{c}{$r=20$}  \\ \cline{2-13} 
         & Acc & Speed & FLOP & Acc & Speed & FLOP & Acc & Speed & FLOP & Acc & Speed & FLOP  \\  \hline
        Full Model & 85.95 & 97.52 & 61.60 & 85.95 & 97.52 & 61.60 & 85.95 & 97.52 & 61.60 & 85.95 & 97.52 & 61.60  \\ \hdashline[0.5pt/1pt]
        ToMe & 84.22 & 182.05 & 30.99 & 60.65 & 266.32 & 20.90 & 13.52 & 350.20 & 15.86 & 4.97 & 427.64 & 12.86  \\ 
        ToFu AVG & 84.51 & \textbf{182.82} & 30.99 & 65.89 & \textbf{268.02} & 20.90 & 16.91 & \textbf{353.16} & 15.86 & 6.98 & \textbf{432.04} & 12.86  \\ 
        ToFU MLERP & \textbf{84.67} & 181.31 & 30.99 & \textbf{67.38} & 266.53 & 20.90 & \textbf{27.71} & 351.29 & 15.86 & \textbf{12.42} & 429.91 & 12.86 \\ \hline
    \end{tabular}
    \caption{Performance on ImageNet with the ViT-L model, utilizing $d\!=\!6$. Across varying backbones and $r$ values, ToFu MLERP consistently surpasses ToMe. Given that ViT-L comprises 24 layers, its token reduction impact is more pronounced than in ViT-B. This accounts for the sharper performance decline observed with increasing $r$ values.}
    \label{tab:large}
\end{table*}

}

\newcommand{\tablebrute}{
\begin{table}[!ht]
    \centering
    \small
    \begin{tabular}{cccc}
    \hline
        Acc1 & \makecell{Merge\\Configuration} & Num Prune & Num Average \\ \hline
        \textbf{80.17} & \texttt{PPPPPPAAAAAA} & 6 & 6 \\ 
        80.00 & \texttt{PPPPPPPAAAAA} & 7 & 5 \\ 
        79.96 & \texttt{PPPPPAAAAAAA} & 5 & 7 \\ 
        ... & ... & ... & ... \\ 
        77.85 & \texttt{AAAAAAPPPPPP} & 6 & 6 \\ 
        77.74 & \texttt{AAAAAAAAPPPP} & 4 & 8 \\ 
        77.73 & \texttt{AAAAAAAPPPPP} & 5 & 7 \\ \hline
    \end{tabular}
    \caption{Ablation study on hybrid merging strategies: pruned merging (P) and average merging (A) with the ViT-B model. The table evaluates ImageNet Top-1 accuracy across different merge configurations. For instance, the configuration \texttt{PPPPPPAAAAAA} represents $d=6$, where the pruned merging is applied to the first 6 layers and average merging to the subsequent 6 layers. }
    \label{tab:d}
\end{table}

}

\newcommand{\tablethree}{
\begin{table}[!ht]
    \setlength\tabcolsep{1.5pt} % default value: 6pt
    \centering
    \small
    \begin{tabular}{lccccc}
    \hline
    Method & Top1 & FLOP & img/s & \makecell{Additional \\ Parameters}  & \makecell{Batching\\Compatible}\\\hline
    DeiT-S & 79.8 & 4.6 & 930 & NA & NA \\
    A-ViT & 78.6 & 2.9 & - & Yes & No\\
    DynamicViT & 79.3 & 2.9 & 1505 & Yes  & No\\
    SP-ViT & 79.3 & 2.6 & - & Yes  & No\\
    $\text { ToMe }_{r_{13}}^{DeiT}$ & 79.4 & \textbf{2.7} & 1552 & \textbf{No} & \textbf{Yes} \\
    $\text { ToMe }_{r_{13}}^{AugReg}$ & 79.3 & \textbf{2.7} & 1550 & \textbf{No}  & \textbf{Yes}\\
    $\text { ToFu }_{r_{13}}^{AugReg}$ & \textbf{79.6} & \textbf{2.7} & \textbf{1561} & \textbf{No} & \textbf{Yes} \\\hline
    \end{tabular}
    \caption{Comparison with SoTA token reduction methods. ``Top1" is the Top-1 accuracy on the ImageNet validation, while ``FLOP" is measured in GFLOPS.``img/s" represents the inference speed. The column "Additional Parameters" indicates whether the method necessitates training an auxiliary network. "Batching Compatible" shows if the method can reduce tokens with batched inputs.}
    \label{tab:sota}
\end{table}
}

\newcommand{\sdtable}{
\begin{table}[t]
\setlength\tabcolsep{3.5pt} % default value: 6pt
    \small
    \centering
    \begin{tabular}{cccccc}
    \hline
        ~ & FID $\downarrow$ & LPIPS $\downarrow$ & MS-SSIM $\uparrow$ & Time $\downarrow$ & Mem $\downarrow$ \\ \hline
        Full & 0.00 & 0.00 & 1.00 & 4.67 sec/img & 11.50G  \\ 
        ToMe & 15.74 & 0.3133 & 0.7304 & 3.19 sec/img & 9.31G  \\ 
        ToFu & \textbf{14.72} & \textbf{0.2706} & \textbf{0.7618} & \textbf{3.16} sec/img & 9.31G \\ \hline
    \end{tabular}
    \caption{Performance of Token Fusion in conditional image generation using the Stable Diffusion method.}
    \label{tab:sd}
\end{table}
}
\newcommand{\figone}{
\begin{figure}[t]
    \centering
    \includegraphics[width=1.0\linewidth]{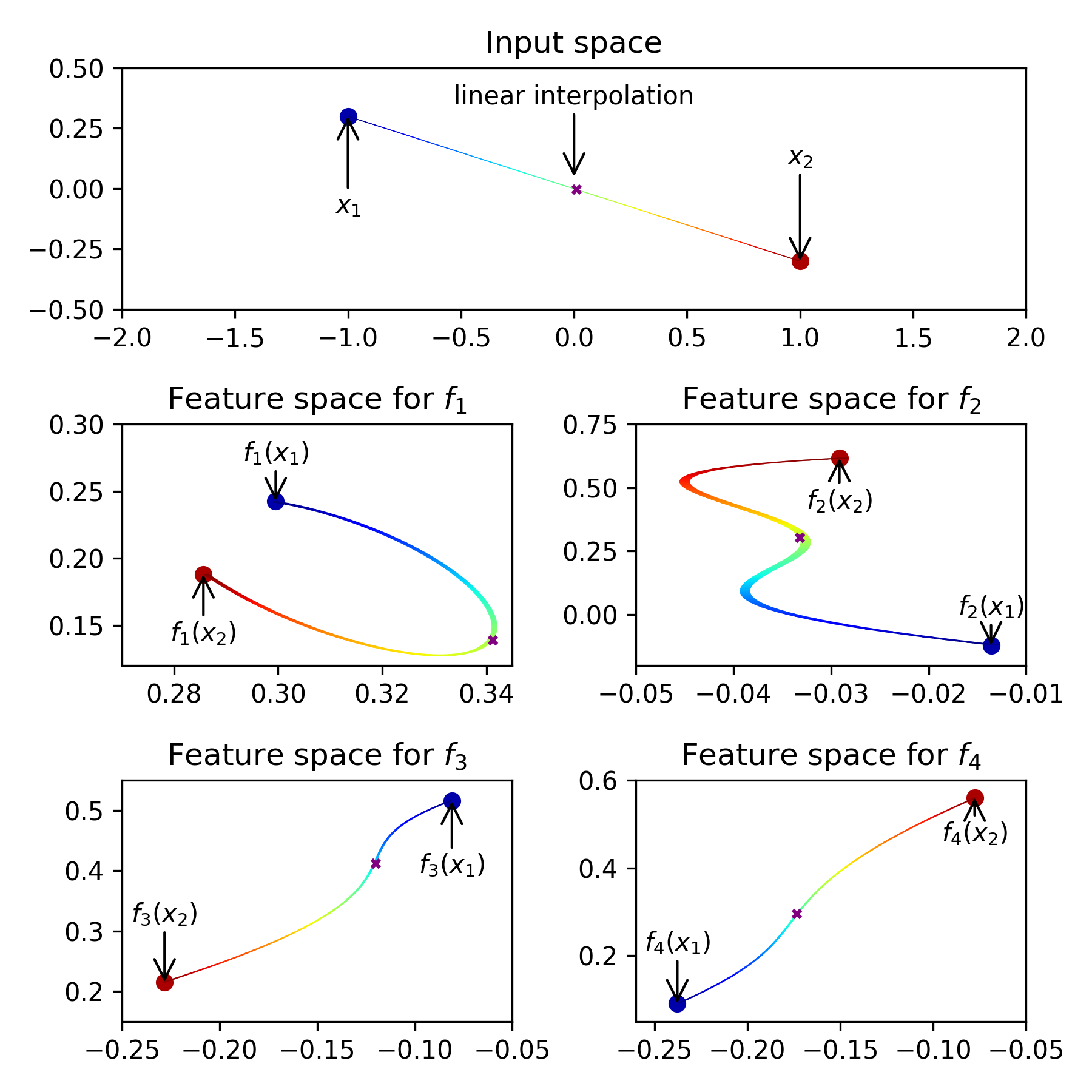}
    \caption{Input interpolation and the corresponding network output of a pretrained ViT across different depths. (Top) Two inputs, $x_1$ and $x_2$, are connected by a linear interpolation, represented by the colored line, with the purple star denoting the average of the two features. (Bottom) Outputs $f_1$ through $f_4$ correspond to the four MLPs from depths $1$ to $4$ of the ViT. Each MLP module is sandwiched between two randomly initialized linear layers: one maps the 2D input to the MLP's dimension, and the other reverts it to 2D for visualization. The output from layers $f_1$ and $f_2$ deviates significantly from the linear interpolation (direct connection) between them. This deviation poses challenges for average merging, potentially producing outputs that diverge from the model's learned distribution.}
    \label{fig:figure1}
\end{figure}
}

\newcommand{\figtwo}{
\begin{figure*}[t]
    \centering
    \includegraphics[width=\linewidth]{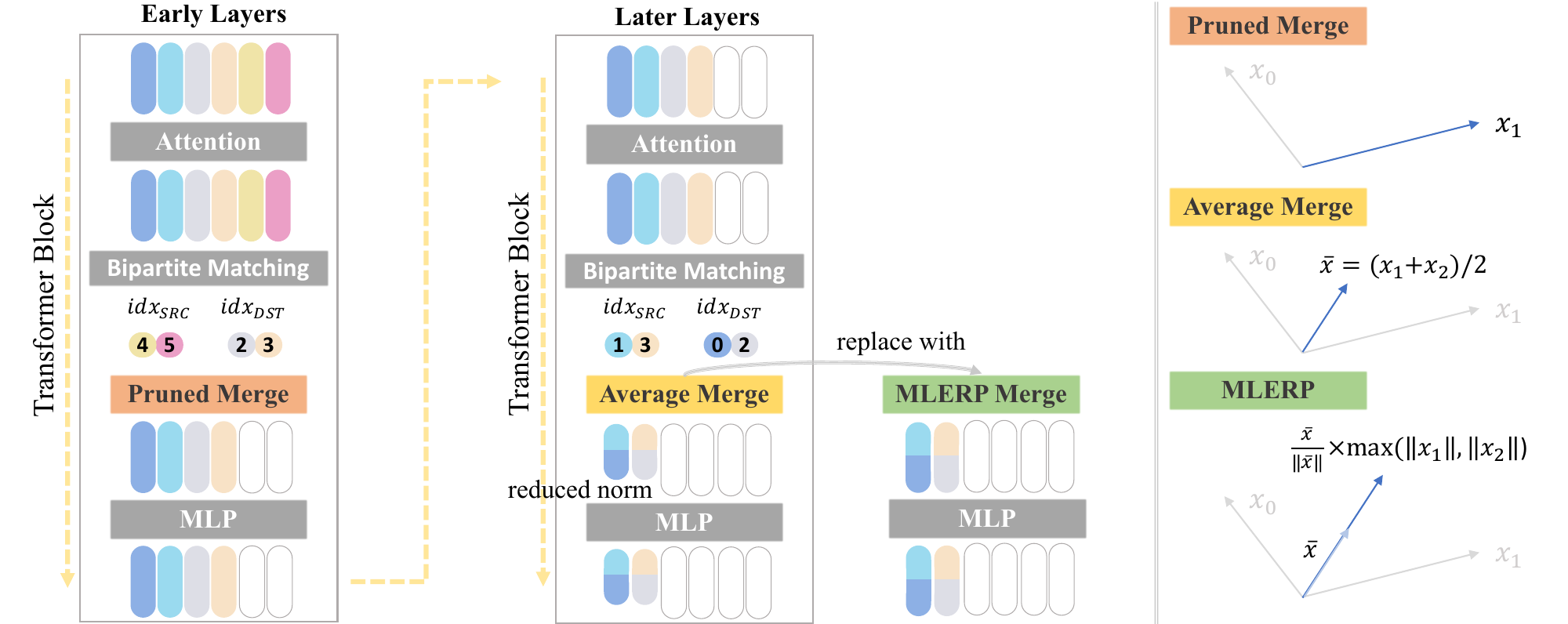}
    \caption{(Left) Token Fusion integration within a ViT backbone. Each Transformer block incorporates a token reduction operation, denoted by \( R \). In this illustration, \( R \) is invoked before the MLP. Using Bipartite Soft Matching (BSM), each \( R \) identifies \( r \) indexes with the highest similarity, for example, \( r=2 \) in this depiction. Notably, the depth of the layer determines the chosen merging strategy: early layers use pruned merging, while later layers transition to average (or MLERP) merging. This dynamic approach accelerates ViT while preserving most of the performance of full token inference. (Right) A visualization of different merging approaches for combining two features, \( x_1 \) and \( x_2 \). All methods can be adapted to merge more than two features. }
    \label{fig:figure2}
\end{figure*}
}

\newcommand{\figlinearity}{
\begin{figure}[t]
    \centering
    \includegraphics[width=\linewidth]{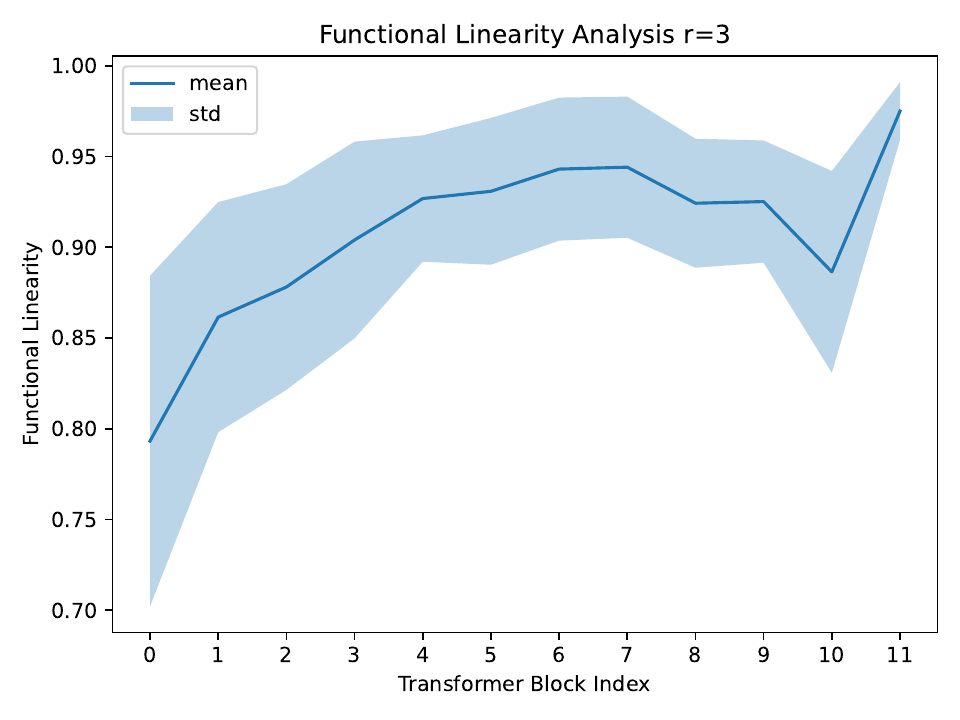}
    \caption{A plot of functional linearity as defined in Eq.~\ref{eq:fl}, measured using a ViT-S pretrained on ImageNet and evaluated with the ImageNet validation set. The interpolation pairs are chosen by the BPE algorithm with \( r=3 \). An increasing trend in functional linearity is also observed in other sizes of ViT. This trend informs our development of a hybrid merging approach in Token Fusion.}
    \label{fig:linearity}
\end{figure}
}

\newcommand{\figtradeoff}{
\begin{figure}[t]
    \centering
    \includegraphics[width=\linewidth]{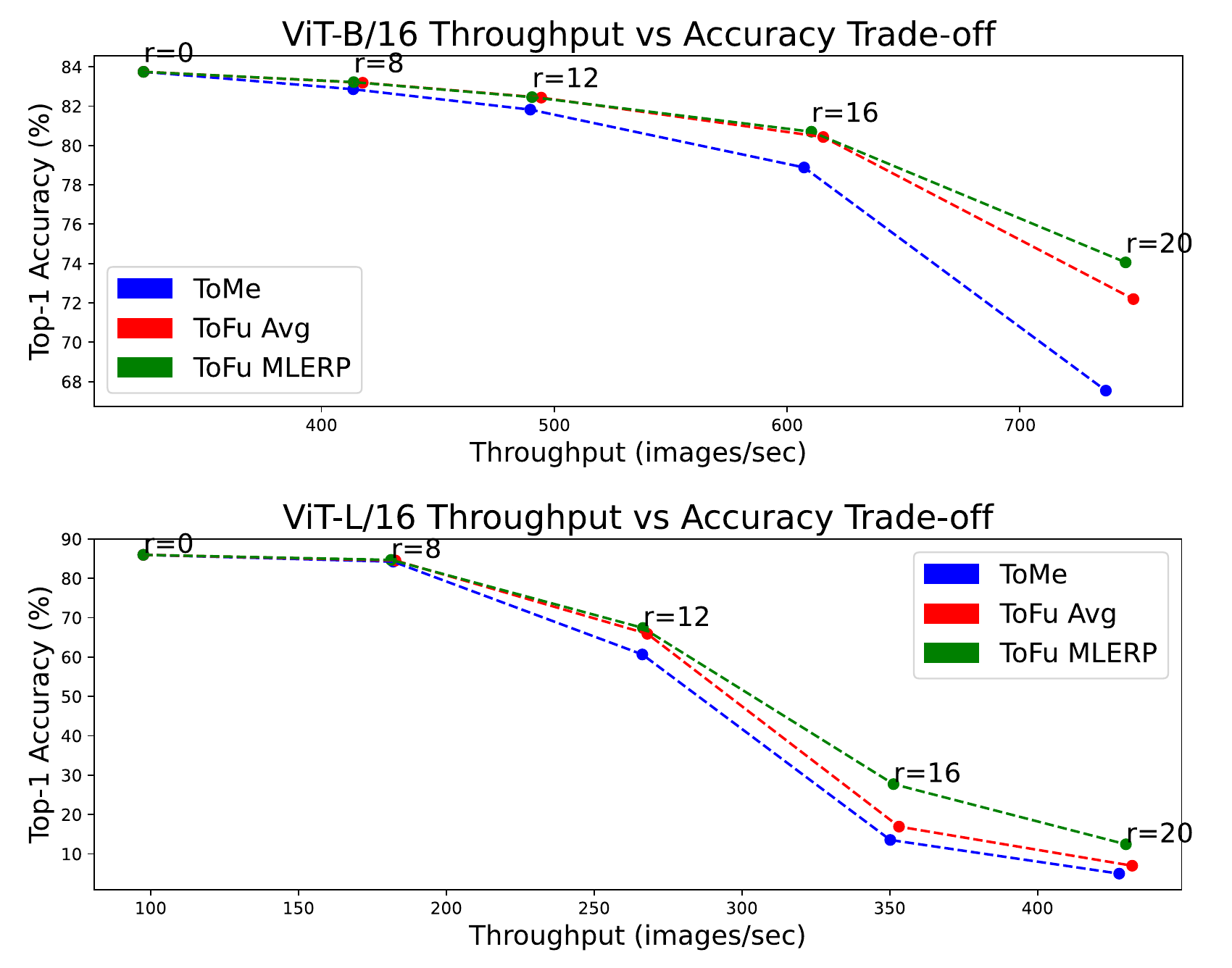}
    \caption{A plot comparing throughput (measured in images/sec) and top1 ImageNet accuracy for ViT-B and ViT-L. The plot shows that ToFu Avg outperforms ToMe in both speed and accuracy. While ToFu MLERP surpasses ToMe in accuracy, its speed is slightly reduced due to norm calculations. Nonetheless, the overall trade-off favors ToFu MLERP as its curve (green) is situated further out compared to ToFu Avg's curve (red).
}
    \label{fig:tradeoff}
\end{figure}
}

\newcommand{\sdimages}{
\begin{figure*}[t]
    \centering
    \includegraphics[width=0.9\linewidth]{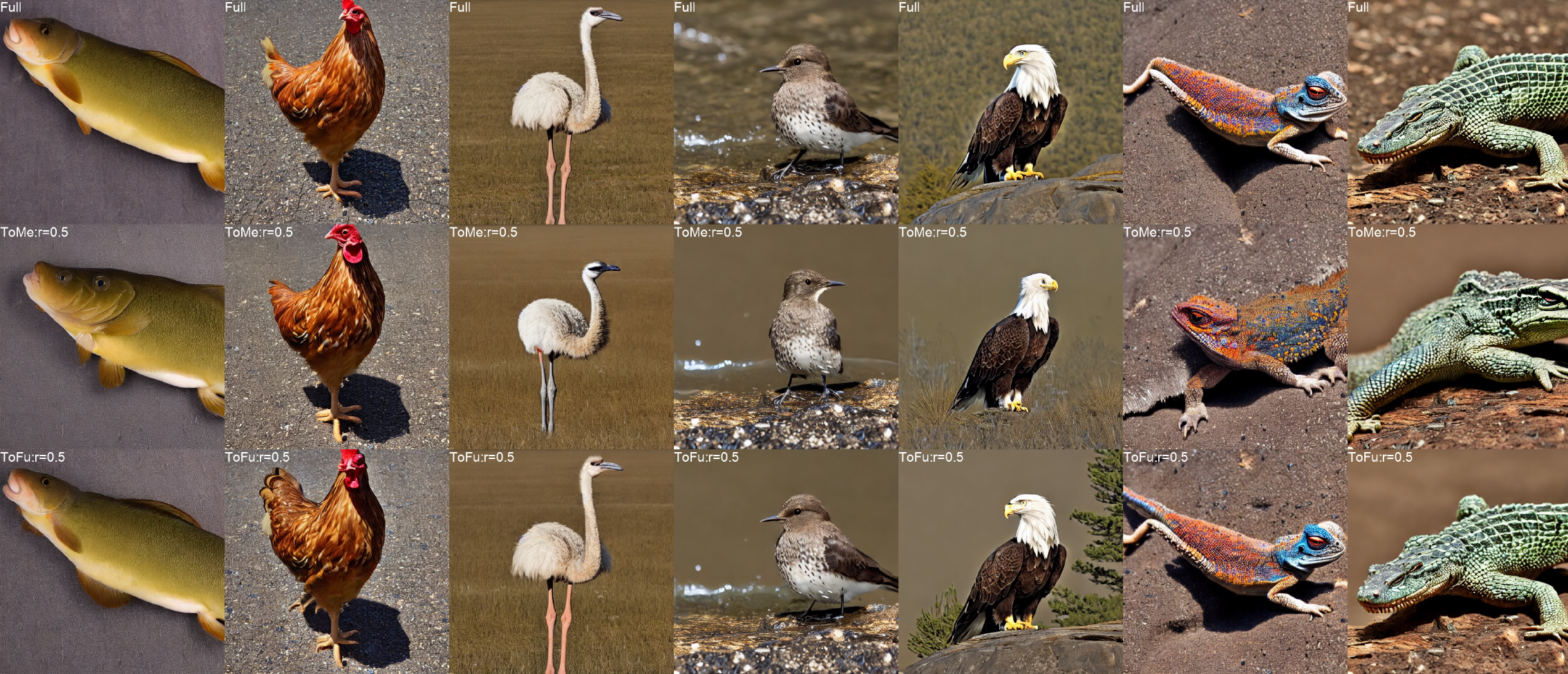}
    \caption{(First row) Images generated with SD v1.5 model using full tokens. (Second row) ToMe: $r=50\%$. (Third row) ToFu: $r=50\%$. The random seed is consistent across each column. The comparison illustrates that ToFu maintains structural consistency in the images more effectively, evident in the consistent head direction pose. However, both ToMe and ToFu compromise background details.  }
    \label{fig:sd}
\end{figure*}
}

% x1+M
% x2+M 

% (x1+x2)/2 + M

% highlight effect

% -cls
% -f1
% -f2
% -f3
% -f4

% -cls
% -f1'
% -f2'

% -cls
% -f1'
% -f2'
% -f3'
% -f4'

% figures 

% x1
% x2
% a
% f(ax1+(1-a)x2)

% f: MLP, ATTN
% Layer 1
% a1 a2   a3 a4 a5
% 0  0.2 0.5 ..    1
% f(a*x2 + (1-a)x1) 
% choose      avg
% f1=f(x1)  f2  f3  f4  f5=f(x2)
% 512      512 512 512 512
%    cos(f2-f1 , f3-f2)

% Layer N
% choose     avg
% f(x1)  f2  f3  f4  f(x2)
% 512    512 512 512 512

% 1   1   1   1 

% continuity of f around x1?
% (x1 + e) -> f(x1+e)
% (x1    ) -> f(x1  )

% x1+x2+x3
% -------
% 3

% vs

% x3 

% x1 vs x2
% f()

% ClS(ViT(X, choose vs avg (interpolate))

\title{Token Fusion: Bridging the Gap between Token Pruning and Token Merging.}

\author{Minchul Kim\thanks{Work completed during an internship at Samsung Research America}\\
% Michigan State University\\
% East Lansing, MI 48824\\
{\tt\small kimminc2@msu.edu}
\and
Shangqian Gao\\
% Samsung Research America\\
% Mountain View, CA, 94043\\
{\tt\small s.gao1@samsung.com}
\and
Yen-Chang Hsu\\
% Samsung Research America\\
% Mountain View, CA, 94043\\
{\tt\small yenchang.hsu@samsung.com}
\and
Yilin Shen\\
% Samsung Research America\\
% Mountain View, CA, 94043\\
{\tt\small yilin.shen@samsung.com}
\and
Hongxia Jin\\
% Samsung Research America\\
% Mountain View, CA, 94043\\
{\tt\small Hongxia.jin@samsung.com}
}

\maketitle

%%%%%%%%% ABSTRACT
\begin{abstract}
Vision Transformers (ViTs) have emerged as powerful backbones in computer vision, outperforming many traditional CNNs. However, their computational overhead, largely attributed to the self-attention mechanism, makes deployment on resource-constrained edge devices challenging. Multiple solutions rely on token pruning or token merging. In this paper, we introduce``Token Fusion" (ToFu), a method that amalgamates the benefits of both token pruning and token merging. Token pruning proves advantageous when the model exhibits sensitivity to input interpolations, while token merging is effective when the model manifests close to linear responses to inputs. We combine this to propose a new scheme called Token Fusion. Moreover, we tackle the limitations of average merging, which doesn't preserve the intrinsic feature norm, resulting in distributional shifts. To mitigate this, we introduce MLERP merging, a variant of the SLERP technique, tailored to merge multiple tokens while maintaining the norm distribution. ToFu is versatile, applicable to ViTs with or without additional training. Our empirical evaluations indicate that ToFu establishes new benchmarks in both classification and image generation tasks concerning computational efficiency and model accuracy.

\end{abstract}

% body
\section{Introduction}

\figone

The recent rise of Vision Transformers (ViT) in the domain of computer vision has undoubtedly marked a paradigm shift from traditional Convolutional Neural Networks (CNNs)~\cite{Mehta2021-fz,Huang2022-pb,Feichtenhofer2022-il,He2021-zh,Graham2021-ch,Fan2021-jj,Dong2021-ay,Liu2021-nh,Dosovitskiy2020-cs,Vaswani2017-qt}. Transformers offer flexibility by allowing long-range interactions between distinct parts of an image. However, with increased power comes increased cost. In this context, the cost translates to a need for more computational resources as resolution increases, particularly due to the self-attention mechanism inherent in transformers. This heightened resource demand often hinders their deployment in real-world, resource-constrained scenarios, especially on edge devices.

Recognizing the need for more efficient transformers, the research community has taken strides in multiple directions. One notable approach is to reduce the number of tokens, which directly leads to an increase in speed. Token pruning has emerged as a pragmatic strategy, wherein certain tokens, considered less consequential to the overall performance, are systematically omitted~\cite{Yin2022-hr,Rao2021-fm}. This determination is rooted in auxiliary loss functions employed during training, which assess the importance of each token. However, this method has its drawbacks. The need to train the model to determine which tokens to prune compromises its user-friendliness. Moreover, pruning a significant number of tokens inevitably results in information loss.

In parallel with token pruning, another intriguing strategy, `token merging,' has emerged~\cite{Bolya2023-ne,Bolya2023-yz,Marin2021-ew}. Unlike pruning, which discards certain tokens entirely, merging consolidates them by averaging tokens that are similar. This approach aims to minimize information loss, ensuring that the most informative aspects of similar tokens are retained even as their total count decreases. ToMe~\cite{Bolya2023-ne,Bolya2023-yz} showcases commendable performance across various tasks, from classification to image generation, and offers the added advantage of being applicable to pretrained models without the need for additional training.

With the emergence of ToME, a pivotal question arises: \textit{While average merging excels in many contexts, does it consistently overshadow the benefits of pruning?} Put another way, are there scenarios where pruning might still be as effective as, or even surpass, the simplicity of average merging? This paper delves deeply into this question, exploring the nuances and strengths of each approach.

\figtwo

Amidst the discourse on token pruning and token merging, we present \textit{Token Fusion} (ToFu). ToFu combines the advantages of both strategies, adjusting according to the model's `functional linearity' concerning its input. This term refers to the degree to which a model's output for interpolated inputs aligns with a linear behavior. Given that neural network blocks are inherently non-linear, the model's response to interpolated inputs, \(x_1\) and \(x_2\)—like in the case of average merging—might deviate from the outputs for the individual inputs, \(f(x_1)\) and \(f(x_2)\). This observation is evident in Fig~\ref{fig:figure1}, where the actual MLP layers (layers 1-4) of pretrained ViTs are studied. The 2D projection of an MLP's response to interpolated inputs emphasizes the variability in functional linearity across different layers.

Specifically, through analysis, we find that token `pruning' gains the upper hand over averaging in scenarios where subsequent operations are low in functional linearity. This is attributed to the fact that the interpolation of inputs could lead to misalignment in the output space, leading to potential information loss or distribution shift. On the contrary, pruning directly operates on existing token representations, eliminating the challenge introduced by such interpolations. However, averaging is beneficial when the model shows high functional linearity, because averaging permits the model to aggregate information from multiple tokens, capturing a more comprehensive and nuanced representation. Drawing from these insights, we integrate this rationale into a single unified algorithm.

Since Token Fusion combines both pruning and merging, we seek to further refine the conventional average merging technique. Standard token merging, in its averaging process, fails to preserve the original feature norm. This can potentially alter the feature distribution and adversely affect its performance. In contrast, pruning leaves the feature norms of the remaining tokens intact. To address the shortcomings of average merging, we introduce \textit{MLERP merging}. Inspired by the SLERP technique, this method is tailored to merge multiple tokens while conserving the norm distribution.

We demonstrate that Token Fusion, a dynamic combination of pruning and merging tailored to each layer's properties, excels over standalone token merging or pruning in ImageNet 1K classification~\cite{Deng2009-on}. Specifically, ToFu not only  delivers superior accuracy than ToMe but also operates faster. Furthermore, when applied to stable diffusion image generation tasks~\cite{stablediffusion}, ToFu outperforms ToMe, producing class conditional images more structurally akin to those generated using full number of tokens.

In summary, our paper contributions are 

\begin{itemize}
    \item Introduction of \textit{Token Fusion} (ToFu), a method that combines token pruning and token merging. ToFu dynamically adapts to each layer's properties, ensuring optimal performance based on the model's functional linearity with respect to the interpolation in its input.
    
    \item Presentation of the \textit{MLERP merging} technique, an enhancement over traditional average merging, inspired by the SLERP method. This approach merges tokens while preserving their norm distribution, addressing a significant limitation in traditional token merging.
    
    \item Empirical validation of ToFu's advantage over token merging and pruning in ImageNet 1K classification and image generation task. Our results show dual advantages: heightened accuracy and faster speed.
    
\end{itemize}

\section{Related Works}

\subsection{Efficient Transformer}

The advent of Vision Transformers (ViT) marked a pivotal shift in computer vision, presenting a compelling alternative to traditional Convolutional Neural Networks (CNNs)~\cite{Mehta2021-fz,Huang2022-pb,Feichtenhofer2022-il,He2021-zh,Graham2021-ch,Fan2021-jj,Dong2021-ay,Liu2021-nh,Dosovitskiy2020-cs,Vaswani2017-qt}. By fostering holistic interactions between image positions through transformer blocks, ViTs addressed the inherent spatial constraints of CNNs. The significant computational cost of the attention modules has spurred efforts toward deploying more efficient ViT models.

Numerous strategies have been proposed to curb the computational demands of the attention mechanism in transformers~\cite{reformer,smyrf,performerslim,longformer,combiner,Wang2020-tz,Dao2022-iy,Shen2021-ou,Chen2021-sk}. Some noteworthy techniques include approximation with hashing (or sparsity) such as those presented in~\cite{reformer,smyrf}, low-rank approximations like~\cite{performerslim}, and a combination of both sparse and low-rank approximation methods to further enhance performance, as seen in~\cite{longformer,combiner}. Moreover, \cite{Liang2022-ba} emphasized the redundancy present within multi-head attention modules, advocating for the removal of such redundancies for efficiency.

However, many of these adaptations primarily focus on accelerating the attention module, which represents just a fraction of the entire network. In contrast, our method emphasizes reducing the number of tokens, directly increasing processing speed. Furthermore, our technique can be seamlessly integrated with works like \cite{Dao2022-iy} that aim to expedite the computation speed of the attention module.

\subsection{Learned Token Reduction}

Efficiency in token usage within ViTs is often achieved through learned token reduction, a technique that identifies and eliminates redundant tokens. This generally necessitates the training of auxiliary models to rank importance of tokens in the input data \cite{Michel2019-es, Rao2021-fm, Liang2022-ba, Yu2023-nj, Fayyaz2022-fz, Song2022-nl, Kong2021-vu, Lassance2021-rh, Kim2022-lk, Kim2020-xb, Goyal2020-ov, Voita2019-qn, Meng2022-jl, Pan2021-ck}. For instance, DynamicViT \cite{Rao2021-fm} employs an MLP to learn pruning masks, further refined with distillation. A-ViT \cite{Yin2022-hr} introduces an efficient approach to ascertain halting probabilities using the first channel of features, complemented by auxiliary losses. Recently, GQA \cite{Ainslie2023-bq} ingeniously proposes sharing key and value heads across groups of query heads, bridging multi-head and multi-query attention.

However, the reliance on auxiliary module fine-tuning is often seen as a drawback, prompting the search for methods that don't require model fine-tuning. Our method, Token Fusion, harmoniously integrates the strengths of both pruning and merging, offering superior performance and efficiency without necessitating auxiliary module fine-tuning, setting it apart from existing techniques.

\subsection{Heuristic Token Reduction}
In contrast to the learned token reduction techniques, several works have proposed heuristic token reduction that can be applied to the off-the-shelf ViTs without further finetuning~\cite{Marin2021-ew,Bolya2023-yz,Bolya2023-ne,Fayyaz2022-fz}. Token Pooling \cite{Marin2021-ew} is one such technique. Adaptive-Token Sampling\cite{Fayyaz2022-fz} samples tokens based on cls token's similarity to other tokens in the attention map, with inverse sampling proving superior to top-k sampling. However, a limitation of this method is its dependence on the class (\texttt{cls}) token which may not be present in dense prediction task such as image generation.

Recently, Token Merging\cite{Bolya2023-ne,Bolya2023-yz} has proposed a novel non-training method which averages similar tokens based on the efficient biparite matching algorithm. In contrast, Token Fusion dynamically synergizes the benefits of pruning and merging, while also addressing the inherent limitations of average merging with  MLERP merging.

\section{Proposed Method}

In this section, we explain the details of Token Fusion. The module is designed to be inserted into the ViT~\cite{Dosovitskiy2020-cs} Transformer blocks. The module increases the model's speed minimal performance trade-offs.

\paragraph{Background}

For a standard ViT, consider an input feature denoted as \( \mathbf{X} \in \mathbb{R}^{N \times C} \). Conventionally, a transformer block processes as follows (omitting LayerNorm for the sake of clarity):
\begin{align}
    \mathbf{X}^* &= \mathbf{X} + \text{ATTN}(\mathbf{X}) \\
    \mathbf{Y} &= \mathbf{X}^* + \text{MLP}(\mathbf{X}^*).
\end{align}
Approaches like token pruning or merging traditionally aim to discover a function \( R \) that reduces the number of tokens to gain computational efficiency. Specifically:
\begin{align}
    \mathbf{X}^* &= R_A(\mathbf{X} + \text{ATTN}(\mathbf{X})) \\
    \mathbf{Y} &= R_M(\mathbf{X}^* + \text{MLP}(\mathbf{X}^*)).
\end{align}
Here, \( R_A \) and \( R_M \)  transform from \( \mathbb{R}^{N \times C} \) to \( \mathbb{R}^{(N-r) \times C} \), when \( r \geq 0 \) for ATTN and MLP. For $r=0$, \( R_A \) and \( R_M \) would simply function as identity operations and for  $r>0$, the output \( \mathbf{Y} \) always retains a smaller number of tokens.

Token Fusion is a generalization of pruning and merging in a unified algorithm. Central to its operation is the Bipartite Soft Matching (BSM) algorithm~\cite{Bolya2023-ne}, which discerns the top $r$ token pairs based on their mutual similarity. Specifically, given two disjoint sets of tokens (SRC, DST), BSM establishes edges between tokens in SRC and DST, assigning weights based on token similarities. The algorithm then selects the top $r$ most similar pairs. Selected pairs are identified by their respective indexes $\text{idx}_{\text{SRC}}$ and $\text{idx}_{\text{DST}}$. 
Upon selection, $r\times 2$ tokens are gathered by \( \text{idx}_{\text{SRC}} \) and \( \text{idx}_{\text{DST}} \) from SRC and DST respectively. While every \( \text{idx}_{\text{SRC}} \) remains unique, the \( \text{idx}_{\text{DST}} \) can recur, as multiple tokens from SRC can be paired with the same token in DST.

%%%%%%%%%%%%%%%%

\subsection{Token Fusion}
\label{sec:R}
Given $r$ similar tokens \( \text{idx}_{\text{SRC}} \) and \( \text{idx}_{\text{DST}} \), from BSM algorithm, we introduce three merging strategies that can be harnessed to reduce the number of tokens.

\paragraph{Average Merging}
The most straightforward approach to fuse tokens is through the average merging strategy.  By averaging, it seamlessly blends information from different tokens, ensuring that no significant features are lost during the merging process. First, the candidate tokens for merging, denoted as `src', are extracted from the SRC based on the \( \text{idx}_{\text{src}} \). This is represented as:
    \begin{equation}
    \text{src} \leftarrow \text{SRC.gather}( \text{idx}_{\text{src}})
    \label{eq:gather}
    \end{equation}
Subsequently, these `src' tokens are averaged into the DST at their designated locations specified by \( \text{idx}_{\text{DST}} \). This is efficiently implemented using the scatter reduce operation:
    \begin{equation}
    \text{dst} \leftarrow \text{DST.scatter\_reduce}(\text{src}, \text{idx}_{\text{dst}}, \text{mode=``mean"})
    \label{eq:scatter}
    \end{equation}
Mode=``mean" refers to averaging the scattered tokens at each index including the DST tokens.

\paragraph{Pruned Merging}

Pruning refers to the elimination of tokens based on specified criteria. While traditional pruning methods target tokens with low feature norms or employ auxiliary models to predict token importance, our approach employs similarity-based pruning, aligned with the principles of average merging. The tokens indexed by \( \text{idx}_{\text{src}} \) and \( \text{idx}_{\text{dst}} \) are deemed to have high similarity. Our strategy pivots on this observation by opting to discard the `src' tokens, premised on the understanding that redundant information can be safely pruned if a similar representation exists in another token. From the implementation perspective, this approach circumvents the operations detailed in Eq.~\ref{eq:gather} and~\ref{eq:scatter}, thus offering faster execution than average merging.

\paragraph{Hybrid Merging} Average merging operates on the presumption that the subsequent non-linear operations (be it MLP or ATTN) maintain a certain `functional linearity' with respect to their inputs. This means that even if we interpolate between two inputs, the output should lie close to the line joining the outputs corresponding to the two original inputs. While complete adherence to this linearity isn't imperative, significant deviations can lead to complications, as illustrated in Fig~\ref{fig:figure1}, where the subsequent operation might stray from the anticipated trajectory, potentially veering away from the known distribution.

In Sec.~\ref{sec:functionallinearityanalysis}, we delve deeper into this property within the ViT model, revealing that its early layers tend to exhibit reduced functional linearity. To exploit this characteristic, our strategy involves leveraging pruned merging in these early layers, which is not affected by functional nonlinearity. Conversely, as we navigate to the deeper layers that retain a more linear functional behavior, we shift to the average merging technique. This dual-approach, aptly termed as \textit{hybrid merging}, ensures that we harness the strengths of both merging methods at different stages of the model. With a hyper-parameter $d$ which can be between $1$ and maximum depth of the model $L$, we let for each layer $l$  
\begin{equation}
    \text{method} = \text{PRUNE } \text{ if } l < d \text{ else } \text{AVG}
\end{equation}
As evidenced in Sec.~\ref{sec:exp}, the decision to employ pruned merging in the early layers (for instance, when \(d=6\)) and to transition to average in subsequent layers considerably bolsters performance.

\paragraph{Functional Linearity Analysis}
\label{sec:functionallinearityanalysis}
To evaluate the functional linearity of various layers in a pretrained ViT, we propose a metric that gauges the ratio of the direct distance between outputs to the distance along the interpolated path. 
Given a function \( f: \mathbb{R}^{C} \rightarrow \mathbb{R}^{C} \) and two points \( X_1,X_2 \in \mathbb{R}^C\), we want to compute the average finite difference between outputs to approximatethe path from $f(X_1)$ to $f(X_2)$. Suppose the changes in $X$ be
\[ X(t) = (1 - t) X_1 + t X_2 \]
where \( t \) is a scalar that varies from 0 to 1. And the magnitude of the change in the function is 
\[ \Delta f(t) = \| f(X(t + \Delta t)) - f(X(t)) \|_2 \]
where \( \Delta t \) is a small change in \( t \), \textit{i.e)} 
$\Delta t = \frac{1}{N-1}$ where $N$ is the number of points in steps $t$. Then the summation of the change is $\sum_{i=1}^{N-1} \Delta f(t_i) $. The Functional Linearity (FL) is
\begin{equation}
\text{FL}(f, X_1, X_2) = \frac{\| f(X_1) - f(X_2) \|_2}{\sum_{i=1}^{N-1} \Delta f(t_i) }
    \label{eq:fl}
\end{equation}
where \( t_i \) is the value of \( t \) at step \( i \). The more linear the function, the higher the value will be. Perfect linearity is $1$.   

\figlinearity
In Fig~\ref{fig:linearity}, we present the functional linearity of a pretrained ViT-S model, evaluated over 50,000 ImageNet images~\cite{Deng2009-on}. The pairs $X_1$ and $X_2$ are taken from the BSM algorithm with $r=5$ and $f$ is set to be the MLP layer of each Transformer Block.  An ascending trend is evident in the initial layers, and notably, as we approach the terminal layers, the FL value converges remarkably close to 1.0. We devise our \textit{hybrid merging} using these characteristics which can be found in various ViT sizes.

\paragraph{MLERP (Norm Preserving Average)} Another crucial observation is that average merging diminishes the feature norm, \textit{i.e.,} \( ||\frac{(x_1 + x_2) }{2}|| \le \max(||x_1||, ||x_2||) \). Given the pivotal role of feature statistics in domain adaptation~\cite{Li2017-wb} and style transfer~\cite{Karras2021-dh,Karras2019-tu}, a reduced feature norm indicates a deviation from the training distribution. A potential correction for two-token merging is to utilize spherical linear interpolation (SLERP)~\cite{slerp}. But, SLERP is structured for 2 samples, rendering it inapplicable for more than two samples. Thus, we introduce a norm-preserving interpolation named Maximum Norm Linear Interpolation (MLERP). For a token set \( \{x_1...,x_K\} \), MLERP is expressed as:
\begin{align}
\bar{x} &= \frac{1}{K}\sum_{k=1}^{K} x_k \\
\text{MLERP}(\{x_1\!...x_K\}) &= \frac{\bar{x}}{\Vert\bar{x}\Vert} \times \max_{k=1,...,K} ||x_k||.
\end{align} 
MLERP computes the normalized average of the features, then scales this normalized average using the maximum norm of the individual vectors. Transitioning from simple average merging to MLERP merging has demonstrably enhanced performance, as illustrated in Tab~\ref{tab:base} and ~\ref{tab:large}.

In summary, the Token Highway function \( R(X, r, \text{method}) \) is outlined in Algorithm~\ref{algo:1} using PyTorch pseudo-code. Here, \( r \) represents the number of tokens to reduce, and \(\text{method}\) can be one of (pruned, average, or MLERP). For hybrid merging, \(\text{method} = \text{pruned}\) is used for layer depth less than 6; otherwise, MLERP is used.

\begin{algorithm}[t]
\caption{Token Fusion}
\begin{algorithmic}[1]
\State \textbf{Input:} \( \mathbf{X}, r, \text{method} \)
\State \textbf{Output:} \( \mathbf{X}_{reduced}, \text{idx}_{\text{src}}, \text{idx}_{\text{dst}} \) 
\State \( B, N, C \leftarrow \text{shape of } x \)
\State \( \text{SRC, DST}, \text{idx}_{\text{SRC}}, \text{idx}_{\text{DST}} \leftarrow \text{partition}(\mathbf{X}) \)
\State \(\text{idx}_{\text{src}}, \text{idx}_{\text{dst}} \leftarrow \text{bipartite\_soft\_matching}(\text{SRC}, \text{DST}, r) \)
\State  \( \text{idx}_{\text{unchanged}} = \text{idx}_{\text{SRC}} \setminus \text{idx}_{\text{src}} \)
\State \( \text{unc} \leftarrow \text{SRC.gather}( \text{idx}_{\text{unchanged}}) \)

\If{\text{method} == \text{`pruned'}}
    
    \State \( \text{dst} = \text{DST} \) \Comment{Not using $\text{idx}_\text{src}$}
    
\ElsIf{\text{method} == \text{`average'}}
    \State \( \text{src} \leftarrow \text{SRC.gather}( \text{idx}_{\text{src}})\)
    \State \( \text{dst} \leftarrow \text{DST.scatter\_reduce}(\text{src}, \text{idx}_{\text{dst}}, \text{mode=mean)} \)
\ElsIf{\text{method} == \text{`MLERP'}}
    \State \( \text{src} \leftarrow \text{SRC.gather}( \text{idx}_{\text{src}})\)
    \State \( \text{dst} \leftarrow \text{DST.scatter\_reduce}(\text{src}, \text{idx}_{\text{dst}}, \text{mode=mean)} \)
    \State \( n\!\leftarrow\!\Vert\text{DST}\Vert\text{.scatter\_reduce}(\Vert\text{src}\Vert,\!\text{idx}_{\text{dst}}, \text{mode:max)}\)
    \State \( \text{dst} \leftarrow \frac{\text{dst}}{\Vert \text{dst} \Vert } \times n \)
\EndIf

\State \( \mathbf{X}_{reduced} \leftarrow \text{concatenate}([\text{unc, dst}], \text{dim=1}) \)

\end{algorithmic}
\label{algo:1}
\end{algorithm}

\section{Experiments}
\label{sec:exp}

We evaluate Token Fusion (ToFu) using ImageNet~\cite{Deng2009-on} trained via MAE~\cite{He2021-zh} and AugReg~\cite{steiner2021train} for classification tasks. Additionally, we test ToFu in conditional image generation using Stable Diffusion~\cite{stablediffusion}. We leverage pretrained model weights from the official MAE~\cite{He2021-zh} repository and TIMM~\cite{rw2019timm}, and are tested across multiple ViT sizes.

The placement of ToFu's reduce operation, $R$, mirrors that of Token Merging~\cite{Bolya2023-ne,Bolya2023-yz}. For classification, the settings align with ToMe~\cite{Bolya2023-ne}, with token numbers linearly decaying by factor $r$. We position $R$ before the MLP module, and not before the ATTN layer. The similarity scores in Bipartite Soft Matching derive from the key in the ATTN module. For the Stable Diffusion image generation~\cite{stablediffusion}, we follow ToMeSD~\cite{Bolya2023-yz}, placing the reduce operation before the ATTN layer in the highest resolution Transformer Blocks, using. The similarity scores in Bipartite Soft Matching derive from $\mathbf{X}$ as opposed to the key in ATTN module. 

Our experiments are conducted on the same setting as ToMe. Inference throughput is gauged on V100 GPUs in FP32 mode, while FLOPs are determined using fvcore~\cite{fvcore}. It's worth noting that image generation speed (in imgs/sec) as opposed to the FLOP is influenced by factors including CPU performance and I/O times; hence, we measure speed on our hardware to ensure consistent comparisons.

\subsection{Classification Task}

In the context of classification, we implement Token Fusion with pretrained ViTs, assessing the balance between ImageNet validation top-1 accuracies and the accompanying efficiency improvements. Tab.\ref{tab:base} and \ref{tab:large} provide an ablation study of the components within our proposed module. ``ToFu AVG" denotes Token Fusion using PRUNED merging for layers $l<d$ and AVG merging otherwise. ``ToFu MLERP" refers to MLERP merging for conditions where AVG merging is used. We set the depth hyper-parameter $d$ at 6 for all models. Thus, the contrast between ToMe and ToFu AVG highlights enhancements from hybrid (PRUNED and AVG) merging. Meanwhile, the comparison of ToFu AVG and ToFu MLERP underscores the advantages of substituting simple average merging with maximum norm merging, preventing the norm's diminishment.

\paragraph{Ablation of Proposed Modules}
Tab.~\ref{tab:base} and \ref{tab:large} detail the performance enhancements seen in both ViT-B and ViT-L when pretrained using MAE~\cite{He2021-zh}. For ViT-B, the accuracy for \( r=8 \) improves from ToMe's \( 82.86\% \) to \( 83.19\% \) with ToFu AVG, and \( 83.22\% \) with ToFu MLERP. This is compared to the full token model performance of \( 83.74\% \). Additionally, these enhancements are coupled with a speed advantage over ToMe. The benefits amplify with higher reduction rates: at \( r=20 \), top-1 accuracy jumps from \( 67.54\% \) to \( 74.06\% \), and speed improves from \( 736.91 \) to \( 745.40 \) images per second. This underscores the merits of hybrid merging and the transition from AVG to MLERP merging, both crucial in preserving performance while operating with fewer tokens. Fig.~\ref{fig:tradeoff} illustrates the efficiency-accuracy trade-off for both ViT-B and ViT-L, with ToFu MLERP emerging as the optimal choice, as evidenced by its curve positioning in the upper right quadrant.

\paragraph{Ablation of \(d\)}
Table \ref{tab:d} ablates the influence of the hyper-parameter \(d\), which designates the transition layer between PRUNED merging and AVG merging in ViT-B. By varying configurations from a complete PRUNE setup (denoted as \texttt{PPPPPPPPPPPP}) to an all AVG scenario (notated as \texttt{AAAAAAAAAAAA}), we rank ToFu's performance based on the top-1 accuracy on the ImageNet Validation set. The delineation in performance between the top three and bottom three configurations underscores the consistent advantage of deploying PRUNE merging prior to AVG merging. A direct comparison reveals that beginning with PRUNE and then transitioning to AVG yields an accuracy of \(80.17\%\), whereas the reverse order (initiating with AVG and later shifting to PRUNE) results in a diminished \(77.73\%\). This stark difference indicates the preferability of refraining from interpolations in the input space during the preliminary layers. Our optimal results are derived with \(d=6\), and this configuration is adopted for all following experiments.

\figtradeoff

\tablebaselong
\tablelargelong

\tablebrute

\paragraph{Comparison with SoTA Methods}
Table \ref{tab:sota} presents a comparative analysis of Token Fusion against prior State-of-the-Art techniques in token pruning and token merging within the ViT-S framework. It is evident that ToMe not only surpasses preceding methodologies but also stands out alongside ToFu as the sole models operable without specific training. Furthermore, their compatibility with batched inputs significantly amplifies the module's usability.
\tablethree

\sdimages

\subsection{Image Generation Task}
Token Fusion (ToFu) is further extended to the text-conditional image generation task using the Stable Diffusion (SD) model \cite{stablediffusion}. The architecture of SD is characterized by a U-Net shape, alternating between ResNet and Transformer blocks. Within this design, the Transformer block serves to integrate text-conditional data into the image feature representation.

Following the configuration in \cite{Bolya2023-yz}, ToFu is incorporated before the ATTN layer within the Transformer block, effectuating a 50\% reduction in the token count (i.e., $r=50\%$). Since the subsequent ResNet requires the full number of tokens, tokens are unmerged to their original length, optimizing the computation speed in the ATTN module. However, ResNet blocks and MLP layers remain unchanged.

The primary objective of applying ToFu in SD is to diminish token counts while ensuring performance remains comparable to using full tokens. To evaluate this, $2$ images are sampled for each of the 1000 ImageNet class labels. Each sample is prompted with \emph{A high-quality photograph of a \{CLASS\}}. Pairwise comparisons between ToFu, ToMe, and the full token model are made, maintaining a consistent random seed in each image generation.

To gauge the effectiveness of image generation, three metrics are employed:
(1) FID~\cite{fid}: Assesses the similarity between class-conditional distributions of the full model and reduced.
(2) LPIPS~\cite{lpips}: Evaluates pairwise perceptual likeness.
(3) MS-SSIM~\cite{msssim,ssim}: Measures pairwise structural consistency.
Both LPIPS and MS-SSIM focus on image-level pairwise comparisons, serving to quantify the deviation from the full token with a fixed radom seed.

In Tab.~\ref{tab:sd}, we present a comprehensive performance comparison. The FID scores indicate that ToFu's image distribution aligns more closely with full token images than ToMe does, signifying its superior capacity to maintain the class conditional distribution. Furthermore, LPIPS and MS-SSIM show ToFu's proficiency in upholding both the structural and perceptual qualities inherent to full token images.

From an efficiency standpoint, ToFu exhibits a modest edge over ToMe in image generation speed, boasting an approximate 32\% enhancement when contrasted with the full model. Fig.~\ref{fig:sd} shows a visual comparison of the generated images from the full model, ToMe, and ToFu (with a reduction rate of $r=50\%$). This comparative visualization explains why ToFu outperforms ToMe in three metrics as ToFu maintains the structural content of full token images. Such an accomplishment can be attributed to the early layer Pruned Merging, which outclasses the AVG merging technique. And in UNet, early layer ATTNs are responsible for setting the overall structural content of the image.

\sdtable

\section{Conclusion}

This paper introduces "Token Fusion" (ToFu), an innovative technique that seamlessly integrates the benefits of token pruning and token merging within Vision Transformers. Guided by the understanding that a model's functional linearity varies with respect to input across different depths, we consolidate both methods into a unified framework. Furthermore, we present MLERP merging as a superior alternative to conventional average merging, ensuring feature norm conservation and thus curbing distribution shifts. Empirical evaluations reinforce ToFu's efficacy, especially in classification and image generation scenarios. Given these promising outcomes, we believe ToFu holds significant potential for optimizing the deployment of ViTs on edge devices.

\paragraph{Limitations}
While Token Fusion (ToFu) has demonstrated notable advantages in the context of Vision Transformers, it is not without limitations. The technique's dependence on the hyperparameter \( d \) for determining the layer at which PRUNED merging switches to AVG merging may necessitate additional tuning for diverse datasets or architectures. Additionally, the benefits of ToFu predominantly arise in scenarios where the functional linearity of the model varies across depths. In cases where this variation is minimal or negligible, the hybrid strategy has to be tuned.

\newpage

%%%%%%%%% REFERENCES
{\small
\bibliographystyle{ieee_fullname}
\bibliography{egbib}

\begin{thebibliography}{10}\itemsep=-1pt

\bibitem{Ainslie2023-bq}
Joshua Ainslie, James Lee-Thorp, Michiel de Jong, Yury Zemlyanskiy, Federico
  Lebr{\'o}n, and Sumit Sanghai.
\newblock {GQA}: Training generalized {Multi-Query} transformer models from
  {Multi-Head} checkpoints.
\newblock In {\em arXiv}, May 2023.

\bibitem{longformer}
Iz Beltagy, Matthew~E Peters, and Arman Cohan.
\newblock Longformer: The long-document transformer.
\newblock {\em arXiv preprint arXiv:2004.05150}, 2020.

\bibitem{Bolya2023-ne}
Daniel Bolya, Cheng-Yang Fu, Xiaoliang Dai, Peizhao Zhang, Christoph
  Feichtenhofer, and Judy Hoffman.
\newblock Token merging: Your {ViT} but faster.
\newblock In {\em {ICLR}}, 2023.

\bibitem{Bolya2023-yz}
Daniel Bolya and Judy Hoffman.
\newblock Token merging for fast stable diffusion.
\newblock In {\em {CVPR} Workshop}, pages 4598--4602, Mar. 2023.

\bibitem{Chen2021-sk}
Tianlong Chen, Yu Cheng, Zhe Gan, Lu Yuan, Lei Zhang, and Zhangyang Wang.
\newblock Chasing sparsity in vision transformers: An {End-to-End} exploration.
\newblock In {\em {NeurIPS}}, volume~34, pages 19974--19988, 2021.

\bibitem{reformer}
Krzysztof Choromanski, Valerii Likhosherstov, David Dohan, Xingyou Song,
  Andreea Gane, Tamas Sarlos, Peter Hawkins, Jared Davis, Afroz Mohiuddin,
  Lukasz Kaiser, et~al.
\newblock Rethinking attention with performers.
\newblock {\em ICML}, 2020.

\bibitem{Dao2022-iy}
Tri Dao, Daniel~Y Fu, Stefano Ermon, Atri Rudra, and Christopher R{\'e}.
\newblock {FlashAttention}: Fast and memory-efficient exact attention with
  {IO-awareness}.
\newblock In S Koyejo, S Mohamed, A Agarwal, D Belgrave, K Cho, and A Oh,
  editors, {\em {NeurIPS}}, volume~35, pages 16344--16359, May 2022.

\bibitem{smyrf}
Giannis Daras, Nikita Kitaev, Augustus Odena, and Alexandros~G Dimakis.
\newblock Smyrf-efficient attention using asymmetric clustering.
\newblock {\em NeurIPS}, 33:6476--6489, 2020.

\bibitem{Deng2009-on}
Jia Deng, Wei Dong, Richard Socher, Li-Jia Li, Kai Li, and Li Fei-Fei.
\newblock {ImageNet}: A large-scale hierarchical image database.
\newblock In {\em {CVPR}}, pages 248--255, June 2009.

\bibitem{Dong2021-ay}
Xiaoyi Dong, Jianmin Bao, Dongdong Chen, Weiming Zhang, Nenghai Yu, Lu Yuan,
  Dong Chen, and Baining Guo.
\newblock {CSWin} transformer: A general vision transformer backbone with
  cross-shaped windows.
\newblock In {\em {CVPR}}, pages 12124--12134, July 2021.

\bibitem{Dosovitskiy2020-cs}
Alexey Dosovitskiy, Lucas Beyer, Alexander Kolesnikov, Dirk Weissenborn,
  Xiaohua Zhai, Thomas Unterthiner, Mostafa Dehghani, Matthias Minderer, Georg
  Heigold, Sylvain Gelly, Jakob Uszkoreit, and Neil Houlsby.
\newblock An image is worth 16x16 words: Transformers for image recognition at
  scale.
\newblock In {\em {ICLR}}, Oct. 2020.

\bibitem{fvcore}
facebookresearch.
\newblock fvcore.
\newblock \url{https://github.com/facebookresearch/fvcore}, 2023.

\bibitem{Fan2021-jj}
H Fan, B Xiong, K Mangalam, Y Li, Z Yan, and J Malik.
\newblock Christoph feichtenhofer. multiscale vision transformers.
\newblock 2021.

\bibitem{Fayyaz2022-fz}
Mohsen Fayyaz, Soroush~Abbasi Koohpayegani, Farnoush~Rezaei Jafari, Sunando
  Sengupta, Hamid Reza~Vaezi Joze, Eric Sommerlade, Hamed Pirsiavash, and
  J{\"u}rgen Gall.
\newblock Adaptive token sampling for efficient vision transformers.
\newblock In {\em {ECCV}}, pages 396--414, 2022.

\bibitem{Feichtenhofer2022-il}
Christoph Feichtenhofer, Haoqi Fan, Yanghao Li, and Kaiming He.
\newblock Masked autoencoders as spatiotemporal learners.
\newblock In S Koyejo, S Mohamed, A Agarwal, D Belgrave, K Cho, and A Oh,
  editors, {\em {NeurIPS}}, volume~35, pages 35946--35958, May 2022.

\bibitem{Goyal2020-ov}
Saurabh Goyal, Anamitra~Roy Choudhury, Saurabh Raje, Venkatesan Chakaravarthy,
  Yogish Sabharwal, and Ashish Verma.
\newblock {{P}o{WER}-{BERT}}: Accelerating {BERT} inference via progressive
  word-vector elimination.
\newblock In Hal~Daum{\'e} Iii and Aarti Singh, editors, {\em {ICML}}, volume
  119 of {\em Proceedings of Machine Learning Research}, pages 3690--3699,
  2020.

\bibitem{Graham2021-ch}
Ben Graham, Alaaeldin El-Nouby, Hugo Touvron, Pierre Stock, Armand Joulin,
  Herve Jegou, and Matthijs Douze.
\newblock {LeViT}: A vision transformer in {ConvNet's} clothing for faster
  inference.
\newblock In {\em {ICCV}}, pages 12259--12269, Oct. 2021.

\bibitem{He2021-zh}
Kaiming He, Xinlei Chen, Saining Xie, Yanghao Li, Piotr Doll{\'a}r, and Ross
  Girshick.
\newblock Masked autoencoders are scalable vision learners.
\newblock In {\em {CVPR}}, pages 16000--16009, Nov. 2021.

\bibitem{fid}
Martin Heusel, Hubert Ramsauer, Thomas Unterthiner, Bernhard Nessler, and Sepp
  Hochreiter.
\newblock Gans trained by a two time-scale update rule converge to a local nash
  equilibrium.
\newblock {\em NeurIPS}, 30, 2017.

\bibitem{Huang2022-pb}
Po-Yao Huang, Hu Xu, Juncheng Li, Alexei Baevski, Michael Auli, Wojciech
  Galuba, Florian Metze, and Christoph Feichtenhofer.
\newblock Masked autoencoders that listen.
\newblock In {\em {NeurIPS}}, volume~35, pages 28708--28720, 2022.

\bibitem{Karras2021-dh}
Tero Karras, Miika Aittala, Samuli Laine, Erik H{\"a}rk{\"o}nen, Janne
  Hellsten, Jaakko Lehtinen, and Timo Aila.
\newblock Alias-free generative adversarial networks.
\newblock In {\em {NeurIPS}}, volume~34, pages 852--863, 2021.

\bibitem{Karras2019-tu}
Tero Karras, Samuli Laine, Miika Aittala, Janne Hellsten, Jaakko Lehtinen, and
  Timo Aila.
\newblock Analyzing and improving the image quality of {StyleGAN}.
\newblock In {\em {CVPR}}, pages 8110--8119, Dec. 2019.

\bibitem{Kim2020-xb}
Gyuwan Kim and Kyunghyun Cho.
\newblock {Length-Adaptive} transformer: Train once with length drop, use
  anytime with search.
\newblock Oct. 2020.

\bibitem{Kim2022-lk}
Sehoon Kim, Sheng Shen, David Thorsley, Amir Gholami, Woosuk Kwon, Joseph
  Hassoun, and Kurt Keutzer.
\newblock Learned token pruning for transformers.
\newblock In {\em Proceedings of the 28th {ACM} {SIGKDD} Conference on
  Knowledge Discovery and Data Mining}, KDD '22, pages 784--794, New York, NY,
  USA, Aug. 2022.

\bibitem{Kong2021-vu}
Zhenglun Kong, Peiyan Dong, Xiaolong Ma, Xin Meng, Mengshu Sun, Wei Niu, Xuan
  Shen, Geng Yuan, Bin Ren, Minghai Qin, Hao Tang, and Yanzhi Wang.
\newblock {SPViT}: Enabling faster vision transformers via soft token pruning.
\newblock In {\em arxiv}, Dec. 2021.

\bibitem{Lassance2021-rh}
Carlos Lassance, Maroua Maachou, Joohee Park, and St{\'e}phane Clinchant.
\newblock A study on token pruning for {ColBERT}.
\newblock In {\em arxiv}, Dec. 2021.

\bibitem{Li2017-wb}
Yanghao Li, Naiyan Wang, Jianping Shi, Jiaying Liu, and Xiaodi Hou.
\newblock Revisiting batch normalization for practical domain adaptation.
\newblock In {\em {ICLR}}, Feb. 2017.

\bibitem{Liang2022-ba}
Youwei Liang, Chongjian Ge, Zhan Tong, Yibing Song, Jue Wang, and Pengtao Xie.
\newblock Not all patches are what you need: Expediting vision transformers via
  token reorganizations.
\newblock Feb. 2022.

\bibitem{performerslim}
Valerii Likhosherstov, Krzysztof~M Choromanski, Jared~Quincy Davis, Xingyou
  Song, and Adrian Weller.
\newblock Sub-linear memory: How to make performers slim.
\newblock {\em NeurIPS}, 34:6707--6719, 2021.

\bibitem{Liu2021-nh}
Ze Liu, Yutong Lin, Yue Cao, Han Hu, Yixuan Wei, Zheng Zhang, Stephen Lin, and
  Baining Guo.
\newblock Swin transformer: Hierarchical vision transformer using shifted
  windows.
\newblock In {\em {ICCV}}, pages 10012--10022, Mar. 2021.

\bibitem{Marin2021-ew}
Dmitrii Marin, Jen-Hao~Rick Chang, Anurag Ranjan, Anish Prabhu, Mohammad
  Rastegari, and Oncel Tuzel.
\newblock Token pooling in vision transformers.
\newblock In {\em arxiv}, Oct. 2021.

\bibitem{Mehta2021-fz}
Sachin Mehta and Mohammad Rastegari.
\newblock {MobileViT}: Light-weight, general-purpose, and mobile-friendly
  vision transformer.
\newblock In {\em ICLR}, Oct. 2021.

\bibitem{Meng2022-jl}
Lingchen Meng, Hengduo Li, Bor-Chun Chen, Shiyi Lan, Zuxuan Wu, Yu-Gang Jiang,
  and Ser-Nam Lim.
\newblock {AdaViT}: Adaptive vision transformers for efficient image
  recognition.
\newblock In {\em {CVPR}}, pages 12309--12318, June 2022.

\bibitem{Michel2019-es}
Paul Michel, Omer Levy, and Graham Neubig.
\newblock Are sixteen heads really better than one?
\newblock In {\em {NeurIPS}}, Sept. 2019.

\bibitem{Pan2021-ck}
Bowen Pan, Rameswar Panda, Yifan Jiang, Zhangyang Wang, Rogerio Feris, and Aude
  Oliva.
\newblock {IA-RED}: {Interpretability-Aware} redundancy reduction for vision
  transformers.
\newblock In {\em {NeurIPS}}, volume~34, pages 24898--24911, 2021.

\bibitem{Rao2021-fm}
Yongming Rao, Wenliang Zhao, Benlin Liu, Jiwen Lu, Jie Zhou, and Cho-Jui Hsieh.
\newblock {DynamicViT}: Efficient vision transformers with dynamic token
  sparsification.
\newblock In {\em {NeurIPS}}, Nov. 2021.

\bibitem{combiner}
Hongyu Ren, Hanjun Dai, Zihang Dai, Mengjiao Yang, Jure Leskovec, Dale
  Schuurmans, and Bo Dai.
\newblock Combiner: Full attention transformer with sparse computation cost.
\newblock {\em NeurIPS}, 34:22470--22482, 2021.

\bibitem{stablediffusion}
Robin Rombach, Andreas Blattmann, Dominik Lorenz, Patrick Esser, and Björn
  Ommer.
\newblock High-resolution image synthesis with latent diffusion models, 2021.

\bibitem{Shen2021-ou}
Zhuoran Shen, Mingyuan Zhang, Haiyu Zhao, Shuai Yi, and Hongsheng Li.
\newblock Efficient attention: Attention with linear complexities.
\newblock In {\em {WACV}}, pages 3531--3539, 2021.

\bibitem{slerp}
Ken Shoemake.
\newblock Animating rotation with quaternion curves.
\newblock In {\em SIGGRAPH}, pages 245--254, 1985.

\bibitem{Song2022-nl}
Zhuoran Song, Yihong Xu, Zhezhi He, Li Jiang, Naifeng Jing, and Xiaoyao Liang.
\newblock {CP-ViT}: Cascade vision transformer pruning via progressive sparsity
  prediction.
\newblock Mar. 2022.

\bibitem{steiner2021train}
Andreas Steiner, Alexander Kolesnikov, Xiaohua Zhai, Ross Wightman, Jakob
  Uszkoreit, and Lucas Beyer.
\newblock How to train your vit? data, augmentation, and regularization in
  vision transformers.
\newblock {\em arXiv preprint arXiv:2106.10270}, 2021.

\bibitem{Vaswani2017-qt}
Ashish Vaswani, Noam Shazeer, Niki Parmar, Jakob Uszkoreit, Llion Jones,
  Aidan~N Gomez, {\L}ukasz Kaiser, and Illia Polosukhin.
\newblock Attention is all you need.
\newblock In {\em {NeurIPS}}, volume~30, 2017.

\bibitem{Voita2019-qn}
Elena Voita, David Talbot, Fedor Moiseev, Rico Sennrich, and Ivan Titov.
\newblock Analyzing multi-head self-attention: Specialized heads do the heavy
  lifting, the rest can be pruned.
\newblock May 2019.

\bibitem{Wang2020-tz}
Sinong Wang, Belinda~Z Li, Madian Khabsa, Han Fang, and Hao Ma.
\newblock Linformer: {Self-Attention} with linear complexity.
\newblock In {\em arxiv}, June 2020.

\bibitem{ssim}
Zhou Wang, Alan~C Bovik, Hamid~R Sheikh, and Eero~P Simoncelli.
\newblock Image quality assessment: from error visibility to structural
  similarity.
\newblock {\em IEEE transactions on Image Processing}, 13(4):600--612, 2004.

\bibitem{msssim}
Zhou Wang, Eero~P Simoncelli, and Alan~C Bovik.
\newblock Multiscale structural similarity for image quality assessment.
\newblock In {\em The Thrity-Seventh Asilomar Conference on Signals, Systems \&
  Computers, 2003}, volume~2, pages 1398--1402. Ieee, 2003.

\bibitem{rw2019timm}
Ross Wightman.
\newblock Pytorch image models.
\newblock \url{https://github.com/rwightman/pytorch-image-models}, 2019.

\bibitem{Yin2022-hr}
Hongxu Yin, Arash Vahdat, Jose~M Alvarez, Arun Mallya, Jan Kautz, and Pavlo
  Molchanov.
\newblock {A-ViT}: Adaptive tokens for efficient vision transformer.
\newblock In {\em {CVPR}}, pages 10809--10818, June 2022.

\bibitem{Yu2023-nj}
Hao Yu and Jianxin Wu.
\newblock A unified pruning framework for vision transformers.
\newblock volume~66, pages 1--2, Apr. 2023.

\bibitem{lpips}
Richard Zhang, Phillip Isola, Alexei~A Efros, Eli Shechtman, and Oliver Wang.
\newblock The unreasonable effectiveness of deep features as a perceptual
  metric.
\newblock In {\em CVPR}, pages 586--595, 2018.

\end{thebibliography}
}

\end{document}